\documentclass[letterpaper]{article}
\usepackage[preprint]{aaai2027}
\usepackage[hyphens]{url}
\usepackage{graphicx}
\usepackage{natbib}
\usepackage{caption}
\usepackage{amsmath,amssymb,amsfonts}
\usepackage{algorithm}
\usepackage{algorithmic}
\usepackage{booktabs}
\usepackage{multirow}

\graphicspath{{../}{./}}

\newcommand{\best}[1]{\textbf{#1}}

\title{A-PAIR: A Benchmark and Identity-Consistent Grounding Framework for Air-Ground Cross-View Referring Person Detection}
\author{
Zhoupeng Guo\textsuperscript{\rm 1},
Xinjie Yao\textsuperscript{\rm 2},
Yunqi Zhu\textsuperscript{\rm 3},
Zhihe Fan\textsuperscript{\rm 4},
Siqi Zhao\textsuperscript{\rm 5},\\
Jianjun Chen\textsuperscript{\rm 5},
Yichen Dong\textsuperscript{\rm 5},
Yan Fan\textsuperscript{\rm 6},
Pengfei Zhu\textsuperscript{\rm 5}
}
\affiliations{
\textsuperscript{\rm 1}School of Automation, Southeast University, Nanjing, China\\
\textsuperscript{\rm 2}Faculty of Information Engineering and Automation, Kunming University of Science and Technology, Kunming, China\\
\textsuperscript{\rm 3}School of Computer Science and Engineering, University of New South Wales, Sydney, Australia\\
\textsuperscript{\rm 4}School of Sports Training, Tianjin University of Sport, Tianjin, China\\
\textsuperscript{\rm 5}Tianjin University, Tianjin, China\\
\textsuperscript{\rm 6}National University of Defense Technology, Changsha, China
}

\begin{document}

\maketitle


\begin{abstract}
Air-ground cross-view referring person detection is a necessary component in the language-to-perception-to-control chain of collective embodied intelligence, grounding a language command into the same physical target before ground and aerial agents can coordinate downstream actions. Existing referring expression comprehension and open-vocabulary grounding methods do not jointly account for cross-view identity consistency, making them insufficient for Air-Ground Cross-View Referring Person Detection (AGCV-RPD), which involves similar pedestrian distractors, weak aerial appearance cues, and cross-view identity consistency. To study this problem, we introduce Air-Ground Paired Identity-Aware Referring (A-PAIR), the first comprehensive AGCV-RPD benchmark, containing $22{,}137$ cross-view referring samples. To construct A-PAIR efficiently, we propose Factorized Annotation and Referential Alignment (FARA), a semi-automatic annotation framework that generates factorized referring descriptions and identity-consistency supervision at reduced cost. We propose Identity-Consistent Referring Grounding (ICRG), a framework that combines factorized referential grounding, candidate-completeness supervision, and cross-view consistency calibration for joint air-ground pair selection. ICRG improves ground, aerial, and pair-level detection over strong baselines, increasing pair F1 from $16.65\%$ to $22.28\%$. These results show that AGCV-RPD requires paired detection and identity-consistent reasoning.
\end{abstract}




\section{Introduction}
\label{sec:intro}

\begin{figure}[t]
\centering
\includegraphics[width=0.9\columnwidth]{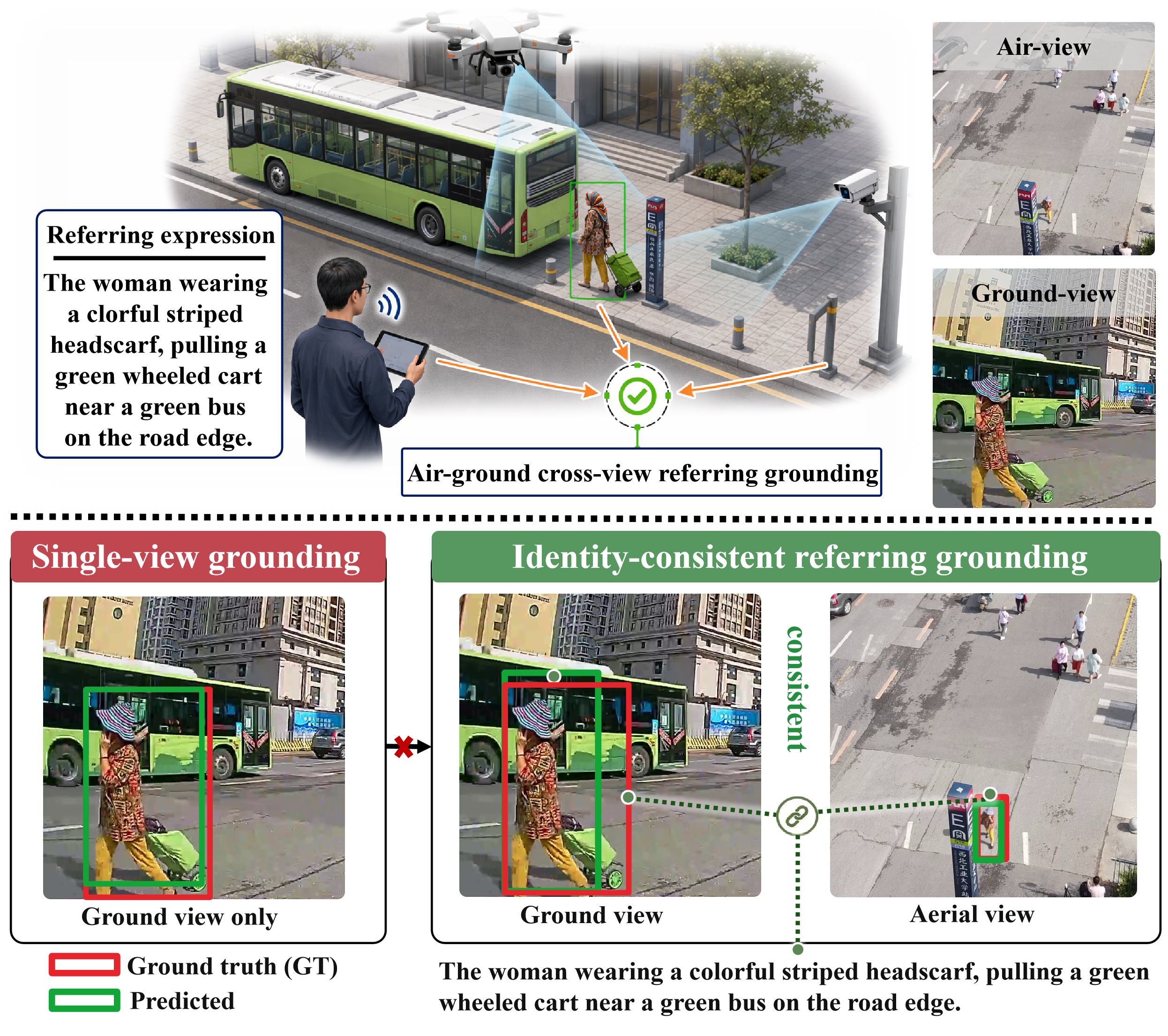}
\caption{AGCV-RPD task. The model detects the same referred person in paired ground and aerial views.}
\label{fig:teaser}
\end{figure}

Air-ground cross-view referring person detection is a necessary component in the language-to-perception-to-control chain of collective embodied intelligence: a language command must first be grounded to the same physical target across complementary sensors before ground and aerial agents can coordinate downstream actions. For such collaborative embodied systems, a fundamental capability is to understand a natural-language referring expression and ground it consistently across heterogeneous sensors. A human or embodied agent may describe a person once, yet expect both the ground and aerial views to detect the same physical individual. We refer to this task as Air-Ground Cross-View Referring Person Detection (AGCV-RPD). Formally, given a ground image, an aerial image of the same scene, and a referring expression, the model must output one bounding box per view that detects the same person identity, as illustrated in Fig.~\ref{fig:teaser}. Solving AGCV-RPD links free-form language, cross-view perception, and identity reasoning into a foundational query interface for air-ground embodied collaboration.

Existing research offers powerful building blocks but does not define the full problem studied here. Single-image referring expression comprehension and open-vocabulary grounding models such as TransVG~\cite{transvg}, MDETR~\cite{mdetr}, GLIP~\cite{glip}, GroundingDINO~\cite{groundingdino}, recent grounding systems~\cite{propvg,sorec,visionr1}, and the drone-oriented RefDrone~\cite{refdrone} provide per-image detection priors for a described object. However, air-ground referring is not a direct extension of single-image grounding. The same scene may contain many similarly dressed pedestrians and repeated spatial layouts, making a referring expression compatible with multiple regions. The aerial view further reduces a person to a small, weakly textured target, so the appearance cues named by language may be barely visible. Most importantly, a plausible prediction in the ground image and a plausible prediction in the aerial image do not necessarily refer to the same physical person. Conversely, a standalone identity matcher assumes candidate crops are already available and cannot decide which regions should be proposed from language. These factors make air-ground referring a coupled detection and identity-consistency problem rather than a pair of independent single-view grounding tasks. This gives rise to three key challenges:
\begin{enumerate}
\item Similar pedestrians and spatial ambiguity.
\item Weak appearance cues of small targets.
\item Cross-view identity consistency.
\end{enumerate}

We address these challenges with Identity-Consistent Referring Grounding (ICRG), an identity-consistent referring grounding framework. Rather than appending a standalone retrieval module after grounding, ICRG builds a unified query-to-pair prediction pipeline around three complementary supervision signals. Factorized referential supervision separates identity-stable appearance semantics from view-dependent scene-spatial semantics to reduce ambiguity among similar pedestrians. Candidate-completeness supervision expands the candidate space so small or weakly described targets are not removed before pair reasoning. Auxiliary cross-view identity-consistency supervision calibrates candidate pairs and suppresses boxes that are individually plausible but jointly identity-inconsistent. To enable this study, we propose Factorized Annotation and Referential Alignment (FARA) to construct Air-Ground Paired Identity-Aware Referring (A-PAIR), a benchmark derived from the G2APS air-ground person dataset with paired images, same-identity boxes, structured text, and auxiliary identity-consistency labels.

Our contributions are as follows:
\begin{itemize}
\item A-PAIR benchmark. We introduce A-PAIR, the first comprehensive benchmark for AGCV-RPD. A-PAIR contains $22{,}137$ cross-view referring samples over $7{,}588$ unique ground images and $3{,}891$ unique aerial images.
\item FARA annotation framework. We propose FARA, a semi-automatic pipeline combining VLM annotation, keyframe propagation, and quality control for cross-view descriptions and identity supervision.
\item ICRG framework. We propose ICRG for the AGCV-RPD task. ICRG integrates candidate-completeness supervision, factorized referential grounding, and cross-view consistency calibration to improve cross-view matching performance in complex air-ground scenes.
\end{itemize}



\section{Related Work}
\label{sec:related}

\subsection{Language-Guided Person Detection}
Language-guided person detection associates textual descriptions with image regions through several complementary research lines. Context-aware methods model attributes, context, and inter-object relations among candidate regions~\cite{mattnet,refcocog,hu_nlobjretrieval}. Transformer approaches, including TransVG~\cite{transvg}, MDETR~\cite{mdetr}, and TransVG++~\cite{transvg_pp}, replace proposal ranking with direct box regression or text-conditioned detection. Region-language pre-training supports open-vocabulary grounding in GLIP~\cite{glip}, GroundingDINO~\cite{groundingdino}, and PropVG~\cite{propvg}. Person-centric research covers text-based and full-image person search~\cite{li_text_person_search,zhang_text_fullimage}, while recent multimodal and small-object methods extend grounding to dialogue, coordinate generation, and tiny targets~\cite{sorec,visionr1}.

Despite these advances, existing methods remain single-view: they predict one-image boxes, retrieve person candidates, or evaluate grounding per image, without requiring ground and aerial predictions to identify the same individual. AGCV-RPD introduces paired air-ground detection, while A-PAIR jointly evaluates language-region alignment and cross-view identity consistency.

\subsection{Cross-View Air-Ground Perception}
Cross-view air-ground perception links observations across viewpoint changes through several complementary research lines. Scene-level geo-localization matches ground images to aerial or satellite views of the same place~\cite{workman_cvusa,hu_cvmnet,zhu_vigor}, learning viewpoint-invariant descriptors across large perspective gaps. Identity-level person re-identification matches people across cameras, from discriminative baselines~\cite{luo_reid_bag} to part-based and transformer representations~\cite{sun_pcb,he_transreid,ye_reid_survey}. Detection-oriented methods use DETR-style and open-vocabulary models for language-aware object detection~\cite{carion_detr,zhu_deformabledetr,minderer_owlvit,zhang_dino,cheng_yoloworld}. Drone-view benchmarks and datasets such as VisDrone and RefDrone~\cite{du_visdrone,refdrone} support person and referring detection in dense high-altitude scenes.

These paradigms nevertheless stop short of paired referring decisions: geo-localization matches places, re-identification assumes given crops, and aerial or open-vocabulary detection remains largely single-view. Drone-view referring adds language guidance but lacks paired ground evidence for identity verification. A-PAIR addresses this gap with paired ground-aerial images, same-identity boxes, auxiliary identity labels, and pair-level evaluation.



\begin{figure*}[t]
\centering
\includegraphics[width=\textwidth]{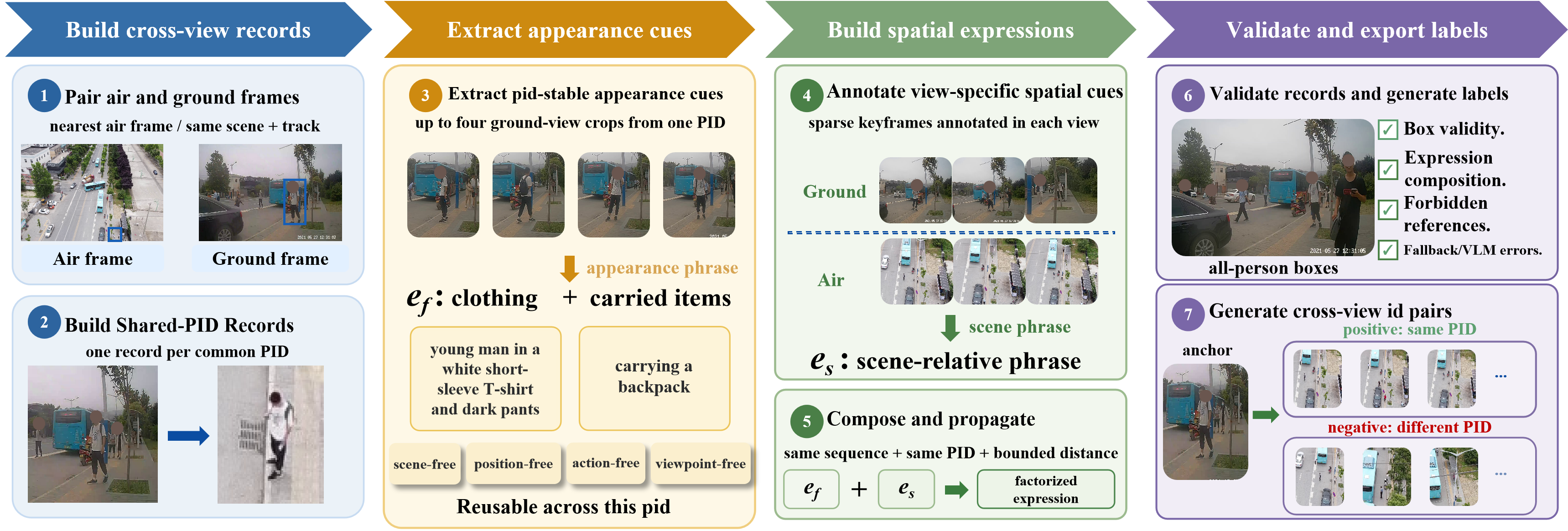}
\caption{FARA annotation pipeline. The pipeline constructs A-PAIR from paired G2APS observations with factorized descriptions and identity labels.}
\label{fig:pipeline}
\end{figure*}

\section{The A-PAIR Benchmark}
\label{sec:dataset}

\subsection{Data Source}
A-PAIR is derived from G2APS~\cite{g2aps}. From paired frames, we retain cross-view observations that share at least one person identity, yielding $7{,}588$ shared-identity ground-aerial pairs associated with $7{,}588$ unique ground images and $3{,}891$ unique aerial images. All bounding boxes follow the $xywh$ convention. Each sample contains both an appearance-full expression and a scene-spatial expression, enabling full-, spatial-, and factorized-prompt evaluation within a unified protocol.

\subsection{FARA annotation}
\label{sec:fara}
The core of A-PAIR is FARA, a semi-automatic factorized annotation framework organized as the seven-step pipeline in Fig.~\ref{fig:pipeline}. Starting from the verified $xywh$ boxes in G2APS Person.mat, FARA groups frames by scene and track band to build cross-view records. Each ground frame is paired with the temporally nearest aerial frame in the same group, and every pid visible in both frames yields one cross-view record with its two target boxes. To extract appearance cues for $e_f$, FARA selects up to four ground-view observations for each pid, preferring distinct images and larger target boxes, and sends padded target crops to a VLM. The VLM returns one reusable noun phrase restricted to stable visible appearance cues (e.g., clothing and clearly carried items); scene context, position, action, viewpoint, and uncertain attributes are excluded. This pid-level phrase is reused by all records of that pid.

FARA independently builds view-specific spatial expressions for the ground and aerial views. It selects sparse keyframes within each view-specific sequence and asks the VLM to assign each target a short scene-relative phrase, such as a relation to a sidewalk, road edge, vehicle, tree, or building. For a non-keyframe observation, FARA copies the nearest keyframe phrase only when it belongs to the same sequence and pid and satisfies both image-distance and frame-gap bounds; otherwise, the observation is marked with a fallback spatial phrase. The final per-view referring expression combines the shared appearance phrase with its view-specific spatial phrase. The released pipeline performs automated, rather than manual, checks of box validity, expression composition, forbidden target/box references, coordinate-only spatial phrases, fallback use, and VLM failures; a completion pass fills missing pid-level appearance profiles.

The remaining steps validate the annotations and export benchmark-ready supervision. FARA augments each record with every Person.mat instance in its two images and exports COCO-style all-person detection labels, providing $32{,}016$ ground and $34{,}108$ aerial training annotations. It also generates cross-view identity pairs: all ground--aerial instance pairs with the same pid are positives, while negatives replace either side with a different-pid person from the same paired images (up to four per replacement direction). The resulting training export contains $377{,}402$ pairs, including $50{,}741$ positives. Finally, FARA forms the official split from connected components of samples that share a ground or aerial image, assigning each whole component to one split.

\subsection{Dataset Statistics and Comparison}

\begin{table*}[t]
\centering
\setlength{\tabcolsep}{2pt}
\begin{tabular*}{\textwidth}{@{\extracolsep{\fill}}lccccccc@{}}
\toprule
Dataset & Lang. & Aerial & Ground & Air-Ground Pair & Same-ID Boxes & ID Aux. & Pair Eval. \\
\midrule
RefCOCO~\cite{refcoco} & \checkmark & - & \checkmark & - & - & - & - \\
RefCOCO+~\cite{refcoco} & \checkmark & - & \checkmark & - & - & - & - \\
RefCOCOg~\cite{refcocog} & \checkmark & - & \checkmark & - & - & - & - \\
RefDrone~\cite{refdrone} & \checkmark & \checkmark & - & - & - & - & - \\
VisDrone~\cite{du_visdrone} & - & \checkmark & - & - & - & - & - \\
UAVDT~\cite{du_uavdt} & - & \checkmark & - & - & - & - & - \\
DOTA~\cite{xia_dota} & - & \checkmark & - & - & - & - & - \\
CVUSA~\cite{workman_cvusa} & - & \checkmark & \checkmark & \checkmark & - & - & - \\
VIGOR~\cite{zhu_vigor} & - & \checkmark & \checkmark & \checkmark & - & - & - \\
Market-1501~\cite{zheng_marketreid} & - & - & \checkmark & - & \checkmark & \checkmark & - \\
DukeMTMC~\cite{li_dukemtmc} & - & - & \checkmark & - & \checkmark & \checkmark & - \\
G2APS~\cite{g2aps} & - & \checkmark & \checkmark & \checkmark & \checkmark & \checkmark & - \\
A-PAIR & \checkmark & \checkmark & \checkmark & \checkmark & \checkmark & \checkmark & \checkmark \\
\bottomrule
\end{tabular*}
\caption{Benchmark comparison. A-PAIR covers language grounding, paired air-ground views, same-identity boxes, and pair-level evaluation.}
\label{tab:benchmark_compare}
\end{table*}

\begin{table}[tbp]
\centering
\setlength{\tabcolsep}{0pt}
\begin{tabular*}{\columnwidth}{@{\extracolsep{\fill}}lcccc@{}}
\toprule
Split & Samples  & Pairs & Gnd.\ img. & Aer.\ img. \\
\midrule
Train & 15{,}497 & 6{,}672 & 6{,}672 & 3{,}530 \\
Val   & 2{,}213  & 279     & 279     & 69     \\
Test  & 4{,}427  & 637     & 637     & 292    \\
\midrule
Total & 22{,}137 & 7{,}588 & 7{,}588 & 3{,}891 \\
\bottomrule
\end{tabular*}
\caption{A-PAIR split statistics.}
\label{tab:dataset}
\end{table}

Table~\ref{tab:dataset} summarizes the split of A-PAIR. It contains $22{,}137$ cross-view referring samples, where each sample is a referring instance with paired ground--aerial images and a target identity; $15{,}497$ are used for training, $2{,}213$ for validation, and $4{,}427$ for testing. The PIDs column counts unique cross-view identities, the Pairs column counts unique ground--aerial image pairs, and the Gnd.\ img. and Aer.\ img. columns count unique ground and aerial images, respectively. 

As shown in Table \ref{tab:benchmark_compare}, the reference, aerial, cross-view, and ReID benchmarks each cover only part of the capabilities of AGCV-RPD. However, they lack paired aerial–ground views, natural-language detection, bounding boxes corresponding to the same identity, or paired evaluation protocols. By providing a dataset that contains language descriptions together with cross-view aerial and ground images, A-PAIR establishes a data foundation for research that bridges the two modalities while maintaining cross-view consistency.

\subsection{Evaluation Metrics}
\label{sec:metrics}
Using the hit definition from the Task Definition subsection, we report instance-level and image-level F1 and accuracy for each view. In this single-target referring protocol, every other person in the image counts as a negative. The instance-level metrics are defined as
\begin{equation}
\text{F1}_{\text{inst}}=\frac{2\text{TP}}{2\text{TP}+\text{FP}+\text{FN}},
\end{equation}
\begin{equation}
\text{Acc}_{\text{inst}}=\frac{\text{TP}}{\text{TP}+\text{FP}+\text{FN}+\text{TN}},
\end{equation}
where $\text{TN}=0$ because there are no no-target samples. Image-level metrics ($\text{F1}_{\text{img}}$, $\text{Acc}_{\text{img}}$) score whether the top prediction attains
\begin{equation}
\mathrm{IoU}(\hat{b}_v,b_v^*) \ge 0.5.
\end{equation}
For pair-level evaluation, a prediction is correct only when both views are correct:
\begin{equation}
\mathrm{hit}_{\text{pair}} = \mathrm{hit}_g \wedge \mathrm{hit}_a.
\end{equation}
The pair metrics (Pair~Acc, Pair~F1) are therefore the primary indicator of cross-view competence.



\section{Method}
\label{sec:method}

\begin{figure*}[t]
\centering
\includegraphics[width=\textwidth]{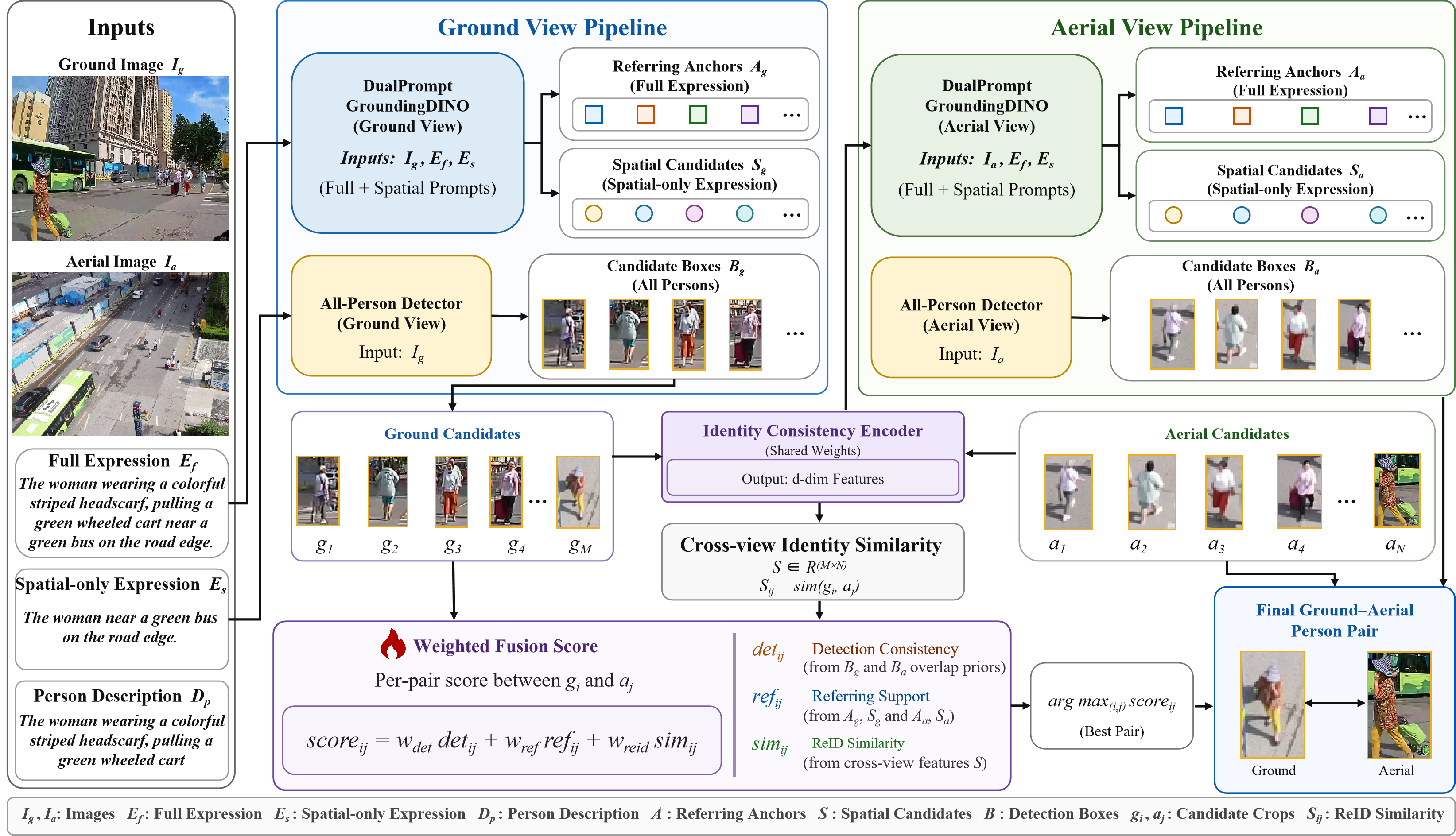}
\caption{ICRG framework. It combines factorized grounding, candidate completion, and cross-view consistency.}
\label{fig:method}
\end{figure*}

ICRG addresses AGCV-RPD with a unified framework for language guided pair detection, as illustrated in Fig.~\ref{fig:method}. Given paired ground and aerial images and factorized referring expressions, it first defines the pair correctness criterion, then builds candidate pools through factorized grounding and all person completion, and finally uses auxiliary identity consistency to calibrate candidate pairs before selecting the final ground and aerial boxes. The following subsections formalize the task, describe the three scoring sources, and present the fusion inference procedure.

\subsection{Task Definition}
\label{sec:task_definition}
Each A-PAIR sample is a tuple $(I_g, I_a, e_f, e_s, y)$, where $I_g$ and $I_a$ are paired ground and aerial images of a shared scene, $e_f$ is an appearance-full referring expression, $e_s$ is a scene-spatial expression, and $y$ is the target person identity (pid). The model must predict one box per view, $(\hat{b}_g, \hat{b}_a)$, both detecting the person with identity $y$. A single-view prediction is correct when its intersection-over-union with the ground-truth box exceeds a threshold $\tau=0.5$:
\begin{equation}
\mathrm{hit}_v = \mathbb{1}\!\left[\mathrm{IoU}(\hat{b}_v, b_v^{*}) \ge \tau\right], \quad v \in \{g, a\}.
\end{equation}
A pair prediction is correct only when both views are correct, $\mathrm{hit}_g \wedge \mathrm{hit}_a$, which makes the pair criterion harder than either view. This coupling is what single-view transfer fails to satisfy and motivates the design below.

\subsection{Factorized Grounding}
The first supervision signal detects candidates referred to by language in each view. Rather than treating a referring expression as an indivisible sentence, A-PAIR factorizes the supervision into an appearance profile for each identity and a scene and spatial description for each view. The full appearance expression $e_f$ captures stable identity semantics, whereas the scene and spatial expression $e_s$ retains spatial cues relative to scene landmarks, such as being near a bus stop shelter, that remain useful when appearance is weak. We feed both factors to a grounding model $G$ built on the GroundingDINO architecture~\cite{groundingdino} with a Swin Transformer image encoder and a text encoder coupled by cross modal attention. For each view $v$ and factor $e \in \{e_f, e_s\}$, the model produces scored referring boxes
\begin{equation}
\mathcal{R}_v^{e} = G(I_v, e) = \{(r_k, s_k^{\mathrm{ref}})\}_{k=1}^{K},
\end{equation}
where $r_k$ is a candidate box and $s_k^{\mathrm{ref}}$ its referring confidence. The grounding model is trained with the standard grounded detection objective, which combines a contrastive token alignment loss and box regression:
\begin{equation}
\mathcal{L}_{\mathrm{ground}} = \mathcal{L}_{\mathrm{align}} + \lambda_{L1}\,\mathcal{L}_{L1} + \lambda_{giou}\,\mathcal{L}_{giou}.
\end{equation}
The appearance factor drives precise detection when appearance is visible, while the scene and spatial factor improves recall in crowded scenes where appearance is ambiguous, providing complementary candidates $\mathcal{R}_v = \mathcal{R}_v^{e_f} \cup \mathcal{R}_v^{e_s}$. This design uses appearance to constrain who the referred person could be and spatial context to constrain where that person can plausibly appear in each view. Taking the union rather than forcing a single expression avoids committing to one cue prematurely and defers the final identity decision to the fusion stage, where evidence from both views is available.

\subsection{Candidate Completeness}
Referring grounding can still miss the target when the expression is under-specified or the target is small. To bound this failure, candidate-completeness supervision trains an all-person detector $D$, sharing the same detector backbone but optimized on the A-PAIR all-person detection labels, to enumerate every person in each view:
\begin{equation}
\mathcal{P}_v = D(I_v) = \{(p_j, s_j^{\mathrm{det}})\}_{j=1}^{M_v},
\end{equation}
with detection confidence $s_j^{\mathrm{det}}$. $D$ is optimized with the detection loss $\mathcal{L}_{\mathrm{det}}=\mathcal{L}_{\mathrm{cls}}+\mathcal{L}_{\mathrm{box}}$. The union $\mathcal{C}_v = \mathcal{R}_v \cup \mathcal{P}_v$ forms the candidate pool per view, so a correct target that the referring signal overlooks can still enter the final selection. The role of $D$ is recall-oriented rather than precision-oriented: because pair selection is decided after cross-view calibration, the detector only needs to place the target somewhere in $\mathcal{C}_v$, and any spurious boxes it introduces are filtered later by referring and consistency scores.

\subsection{Cross-View Consistency}
The third supervision signal calibrates whether two language-compatible candidates also refer to the same physical person. For a candidate box $c$ in view $v$, we crop the region and embed it with a ResNet-50 encoder $f$ into a unit-normalized feature $f(c)\in\mathbb{R}^{d}$. The auxiliary identity-consistency score between a ground candidate $g_i$ and an aerial candidate $a_j$ is the cosine similarity
\begin{equation}
s^{\mathrm{id}}(g_i, a_j) = \frac{f(g_i)^{\top} f(a_j)}{\lVert f(g_i)\rVert\,\lVert f(a_j)\rVert}.
\end{equation}
The encoder is trained on cross-view crop pairs from the training split using the binary consistency loss
\begin{equation}
\begin{aligned}
\mathcal{L}_{\mathrm{id}}
&= -\!\!\sum_{(i,j)\in\mathcal{T}}
\Big[
y_{ij}\log \sigma\!\left(s^{\mathrm{id}}_{ij}\right) \\
&\qquad +(1-y_{ij})
\log\!\left(1-\sigma\!\left(s^{\mathrm{id}}_{ij}\right)\right)
\Big],
\end{aligned}
\label{eq:id}
\end{equation}
where $\mathcal{T}$ is the set of training cross-view crop pairs,
$y_{ij}=\mathbb{1}\!\left[\mathrm{pid}(g_i)=\mathrm{pid}(a_j)\right]$
indicates whether the two crops share the same pid, and $\sigma$ is the logistic function. The signal supplies cross-view identity consistency that language alone cannot provide. It learns to bridge the appearance gap between a street-level crop, where clothing and build are visible, and an aerial crop of the same person, where those attributes are lost to scale and viewpoint. The signal is auxiliary: it does not retrieve a person independently of language, but calibrates pairs proposed by the referring and completeness signals. It can suppress an identity-inconsistent pair in which each view matches the expression but both boxes depict different people.
\subsection{Fusion and Inference}
The final stage selects one identity-consistent ground-aerial candidate pair. For each candidate pair $(g_i, a_j)\in\mathcal{C}_g \times \mathcal{C}_a$ we compute a fused score
\begin{equation}
\mathrm{score}(g_i, a_j) = w_{\mathrm{det}}\, s^{\mathrm{det}}_{ij} + w_{\mathrm{ref}}\, s^{\mathrm{ref}}_{ij} + w_{\mathrm{id}}\, s^{\mathrm{id}}(g_i, a_j),
\label{eq:fusion}
\end{equation}
where $s^{\mathrm{det}}_{ij}$ and $s^{\mathrm{ref}}_{ij}$ aggregate the detection and referring confidences of the two boxes. The non-negative weights $(w_{\mathrm{det}}, w_{\mathrm{ref}}, w_{\mathrm{id}})$ are not learned by gradient descent; they are selected by grid search on the validation split to maximize pair accuracy, which keeps the fusion interpretable and avoids overfitting the small test set. At inference, the predicted pair is the arg-max of Eq.~\eqref{eq:fusion}:
\begin{equation}
(\hat{b}_g, \hat{b}_a) = \arg\max_{(g_i, a_j)} \mathrm{score}(g_i, a_j).
\end{equation}
Algorithm~\ref{alg:infer} summarizes inference. All learnable components ($G$, $D$, $f$) are trained on the training split with their objectives; the fusion weights are fixed once selected on validation, and the test split is evaluated once.

\begin{algorithm}[t]
\caption{ICRG inference. Candidate pairs are scored by fused detection, referring, and identity-consistency cues.}
\label{alg:infer}
\begin{algorithmic}[1]
\REQUIRE ground image $I_g$, aerial image $I_a$, prompts $e_f, e_s$
\STATE $\mathcal{R}_v \leftarrow G(I_v,e_f)\cup G(I_v,e_s)$ for $v\in\{g,a\}$
\STATE $\mathcal{P}_v \leftarrow D(I_v)$;\quad $\mathcal{C}_v \leftarrow \mathcal{R}_v\cup\mathcal{P}_v$
\FORALL{$(g_i,a_j)\in \mathcal{C}_g\times\mathcal{C}_a$}
\STATE compute $s^{\mathrm{det}}_{ij}, s^{\mathrm{ref}}_{ij}, s^{\mathrm{id}}(g_i,a_j)$
\STATE $\mathrm{score}(g_i,a_j)\leftarrow$ Eq.~\eqref{eq:fusion} with val-selected weights
\ENDFOR
\RETURN $\arg\max_{(g_i,a_j)}\mathrm{score}(g_i,a_j)$
\end{algorithmic}
\end{algorithm}



\begin{table*}[tbp]
\centering
{%
\small
\setlength{\tabcolsep}{2pt}
\begin{tabular*}{\textwidth}{@{\extracolsep{\fill}}l cccc cccc cc@{}}
\toprule
\multirow{2}{*}{Method} & \multicolumn{4}{c}{Ground} & \multicolumn{4}{c}{Aerial} & \multicolumn{2}{c}{Pair} \\
\cmidrule(lr){2-5}\cmidrule(lr){6-9}\cmidrule(lr){10-11}
 & F1$_i$ & Acc$_i$ & F1$_{\text{img}}$ & Acc$_{\text{img}}$ & F1$_i$ & Acc$_i$ & F1$_{\text{img}}$ & Acc$_{\text{img}}$ & Acc & F1 \\
\midrule
TransVG~\cite{transvg}               & 2.53 & 1.28 & 4.94 & 2.53 & 0.05 & 0.02 & 0.09 & 0.05 & 0.00 & 0.00 \\
PropVG~\cite{propvg}                 & 1.49 & 0.75 & 2.94 & 1.49 & 0.00 & 0.00 & 0.00 & 0.00 & 0.00 & 0.00 \\
Vision-R1~\cite{visionr1}            & 1.92 & 0.97 & 3.77 & 1.92 & 0.00 & 0.00 & 0.00 & 0.00 & 0.00 & 0.00 \\
SOREC-PIZA~\cite{sorec}              & 13.96 & 7.50 & 24.50 & 13.96 & 7.43 & 3.86 & 13.84 & 7.43 & 1.78 & 3.51 \\
GDINO-B~\cite{groundingdino}         & 26.74 & 15.44 & 42.20 & 26.74 & 12.02 & 6.39 & 21.46 & 12.02 & 5.74 & 10.85 \\
RefDrone~\cite{refdrone}             & 33.50 & 20.12 & 50.19 & 33.50 & 14.66 & 7.91 & 25.57 & 14.66 & 8.31 & 15.35 \\
GDINO-T~\cite{groundingdino}         & 32.28 & 19.25 & 48.80 & 32.28 & 17.42 & 9.54 & 29.67 & 17.42 & 9.08 & 16.65 \\
ICRG (ours)                          & \best{35.17} & \best{21.34} & \best{52.04} & \best{35.17} & \best{21.89} & \best{12.29} & \best{35.92} & \best{21.89} & \best{12.54} & \best{22.28} \\
\bottomrule
\end{tabular*}
}%
\caption{Main results on A-PAIR. All metrics are percentages; bold and italics mark the best and second-best results.}
\label{tab:main}
\end{table*}



\begin{table}[tbp]
\centering
{%
\small
\setlength{\tabcolsep}{0pt}
\begin{tabular*}{\columnwidth}{@{\extracolsep{\fill}}ccc cccccc@{}}
\toprule
Fact. & Cand. & Cons. & \multicolumn{2}{c}{Ground} & \multicolumn{2}{c}{Aerial} & \multicolumn{2}{c}{Pair} \\
\cmidrule(lr){4-5}\cmidrule(lr){6-7}\cmidrule(lr){8-9}
 & & & F1$_i$ & F1$_{\text{img}}$ & F1$_i$ & F1$_{\text{img}}$ & Acc & F1 \\
\midrule
- & - & - & 7.32 & 13.65 & 19.91 & 33.21 & 1.83 & 3.60 \\
\checkmark & - & - & 34.88 & 51.72 & 20.83 & 34.47 & 11.66 & 20.88 \\
\checkmark & \checkmark & - & \best{36.12} & \best{53.07} & 21.82 & 35.82 & 12.24 & 21.82 \\
\checkmark & \checkmark & \checkmark & 35.17 & 52.04 & \best{21.89} & \best{35.92} & \best{12.54} & \best{22.28} \\
\bottomrule
\end{tabular*}
}%
\caption{Ablation results on A-PAIR. Columns indicate enabled supervision signals under the same protocol.}
\label{tab:ablation}
\end{table}



\section{Experiments}
\label{sec:exp}

\subsection{Setup}
Dataset, protocol, and metrics. All experiments use the A-PAIR dataset and the evaluation protocol from Evaluation Metrics, with $15{,}497$ training, $2{,}213$ validation, and $4{,}427$ test samples. Unless noted, methods are trained on the training split and evaluated once on the test split. We report the per-view instance-level and image-level metrics and the pair-level metrics defined in Evaluation Metrics.

\subsection{IoU-Quality Diagnosis Reveals Core Challenges}
\begin{figure}[tbp]
\centering
\includegraphics[width=1.0\columnwidth]{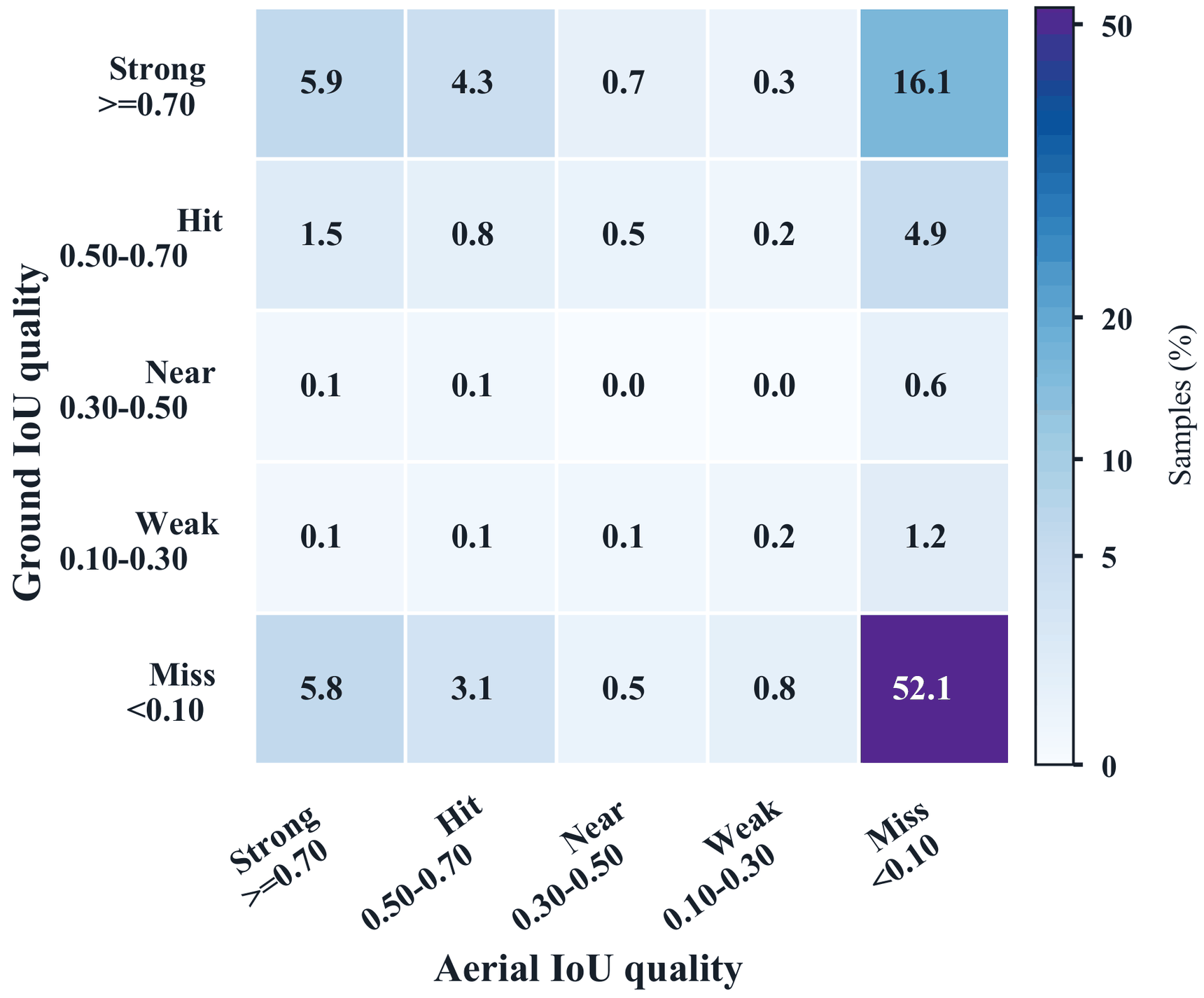}
\caption{Pair-level IoU-quality decomposition. Each cell reports the percentage of A-PAIR test samples in a ground-aerial IoU-quality bin.}
\label{fig:error_decomposition}
\end{figure}

Before comparing methods, we use Fig.~\ref{fig:error_decomposition} to diagnose where AGCV-RPD predictions fail. The cells where both views satisfy $\mathrm{IoU}\ge0.5$ correspond to pair-level hits and account for only $12.54\%$ of the test set, showing that correct paired detection is much harder than obtaining plausible boxes in individual views. The distribution further reveals three recurring failure patterns. First, aerial detection is the main bottleneck: many samples contain a correct or near-correct ground prediction but miss the aerial target, consistent with the weak appearance cues of small aerial persons. Second, the Near and Weak bins indicate that many errors are not arbitrary background selections; the model often predicts nearby or partially overlapping regions, suggesting ambiguity among similar pedestrians and repeated spatial layouts. Third, the gap between view-wise hits and pair-level hits shows that independent single-view compatibility does not reliably produce an identity-consistent pair, motivating explicit cross-view consistency modeling.

\subsection{Main Results Validate Paired Reasoning}
\begin{figure}[tbp]
\centering
\includegraphics[width=0.80\columnwidth]{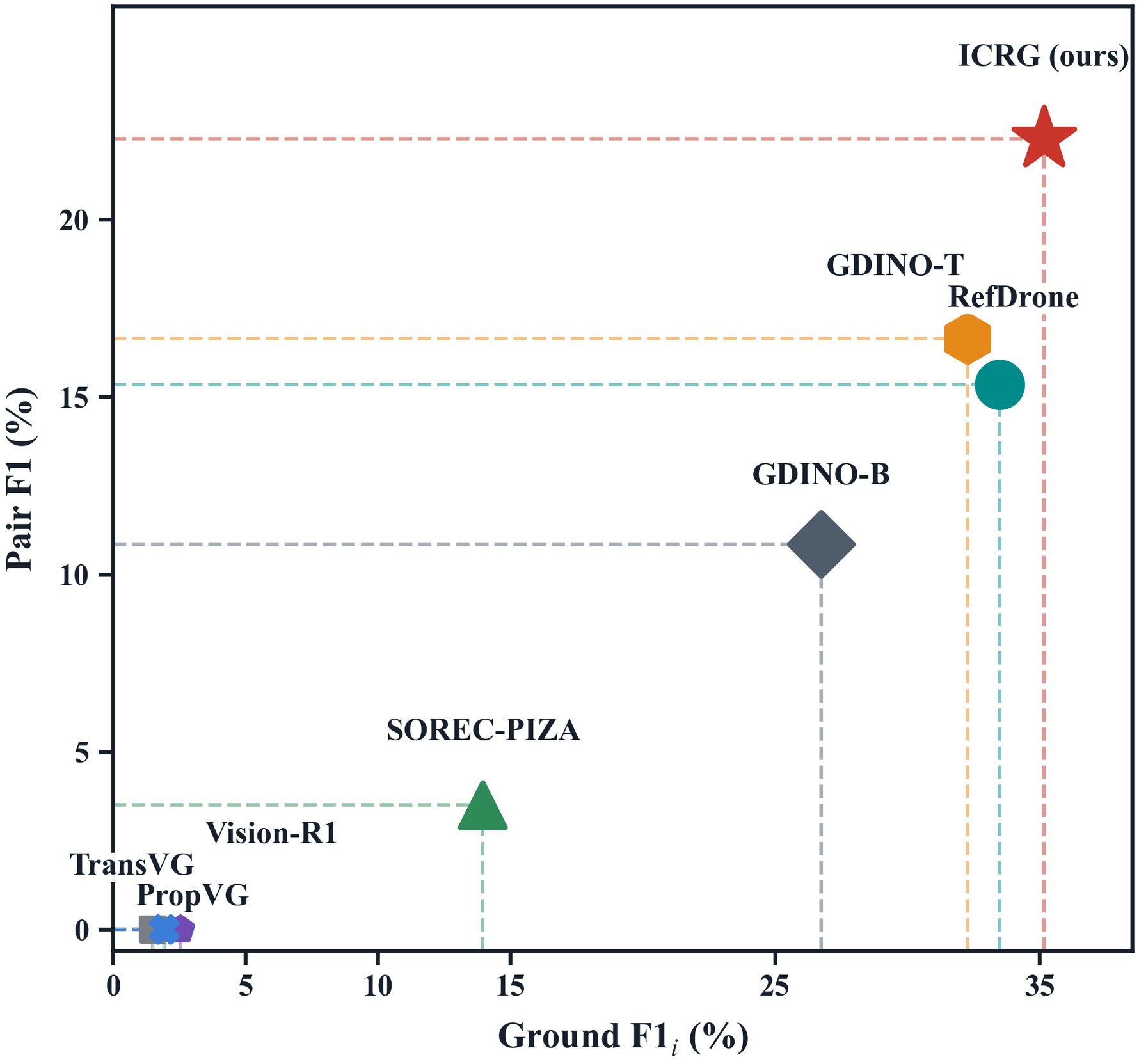}
\caption{Ground versus pair performance. Each point shows one method by Ground F1$_i$ and Pair F1 on A-PAIR.}
\label{fig:ablation}
\end{figure}

Table~\ref{tab:main} compares ICRG with seven single view baselines on the image disjoint test split, including REC, small object grounding, and VLM-based detection models. The strongest baseline, GroundingDINO-T~\cite{groundingdino}, obtains $16.65\%$ pair F1, while ICRG improves it to $22.28\%$. This gain shows that independently grounding the two views is insufficient for AGCV-RPD, because two plausible single view predictions may still fail to form an identity consistent pair. The weak pair results of recent methods such as SOREC-PIZA, PropVG, and Vision-R1 further support this conclusion. As visualized in Fig.~\ref{fig:ablation}, ICRG achieves the best tradeoff between ground view detection and pair level performance, validating the need for joint language grounding, candidate recall, and cross view consistency.

\subsection{Factorized Grounding Reduces Ambiguity}  
Factorized grounding addresses similar pedestrians and spatial ambiguity. In Table~\ref{tab:ablation}, enabling only Fact. improves pair F1 from $3.60\%$ to $20.88\%$. This improvement indicates that separating appearance and spatial cues is more effective than treating the referring expression as an undifferentiated prompt. The appearance factor narrows who the target could be, while the scene and spatial factor narrows where the target can plausibly appear in each view. Their union provides complementary candidates and reduces early mistakes caused by ambiguous descriptions or visually similar pedestrians.

\subsection{Completeness Improves Aerial Recall}
Candidate completeness mainly benefits aerial detection, where targets are small and appearance cues are weak. In Table~\ref{tab:ablation}, adding Cand. to the setting with Fact. alone raises aerial F1$_i$ from $20.83\%$ to $21.82\%$ and aerial F1$_{\text{img}}$ from $34.47\%$ to $35.82\%$, with the full model further reaching $21.89\%$ and $35.92\%$. The larger image-level gain suggests that Cand. recovers valid targets in difficult aerial frames, while the smaller instance-level change indicates that additional proposals may also introduce distractors. The all-person detector therefore serves as a recall safeguard. When the referring branch misses a tiny or under-specified aerial target, candidate completion can still retain that person for later pair selection. By decoupling proposal recall from language confidence, Cand. prevents an early grounding miss from becoming irreversible. Its gain thus comes from improving target availability rather than acting as a standalone precision module.

\subsection{Consistency Calibration Improves Pairing}  
Consistency calibration improves the final pair decision after candidate generation produces plausible boxes in both views. In Table~\ref{tab:ablation}, adding Cons. to the Fact.+Cand. setting raises pair Acc from $12.24\%$ to $12.54\%$ and pair F1 from $21.82\%$ to $22.28\%$. These gains are modest but meaningful because Cons. does not create new candidates; it reranks the existing ground--aerial candidate pairs according to identity compatibility. Thus, identity consistency changes the final decision from selecting two separately plausible boxes to selecting a jointly valid pair, although missed aerial targets and similar distractors still limit absolute pair accuracy.

\section{Conclusion}
\label{sec:conclusion}
We introduced air-ground cross-view referring person detection and developed A-PAIR, a benchmark built from paired ground-aerial observations with same-identity boxes, factorized language expressions, all-person detection labels, and identity-consistency pairs. A-PAIR formulates language-guided person detection as a joint cross-view decision problem: the predicted ground and aerial boxes must both match the query and correspond to the same physical individual. Existing single-view referring methods process the two views independently, making them vulnerable to ambiguous candidates and cross-view identity mismatches. To address this challenge, ICRG unifies factorized referential grounding, candidate-completeness supervision, and cross-view consistency calibration in a query-to-pair framework. On the A-PAIR dataset, ICRG consistently improves ground-view, aerial-view, and pair-level detection over strong single-view baselines, increasing pair F1 from $16.65\%$ to $22.28\%$. These results demonstrate the value of explicitly modeling cross-view identity consistency for paired referring detection. Future work will extend A-PAIR toward temporal air-ground observations, more crowded scenes, and broader environmental diversity.


\bibliography{main}

\end{document}